\documentclass[sigconf]{acmart}
\AtBeginDocument{%
  }

\setcopyright{acmlicensed}
\copyrightyear{2018}
\acmYear{2018}
\acmDOI{XXXXXXX.XXXXXXX}
\acmConference[Conference acronym 'XX]{Make sure to enter the correct
  conference title from your rights confirmation emai}{June 03--05,
  2018}{Woodstock, NY}
\acmISBN{978-1-4503-XXXX-X/18/06}
\usepackage{bm}
\usepackage{colortbl,xcolor}
\usepackage{xspace}
\usepackage{footnote}
\makeatletter
\DeclareRobustCommand\onedot{\futurelet\@let@token\@onedot}
\def\@onedot{\ifx\@let@token.\else.\null\fi\xspace}

\def\eg{\emph{e.g}\onedot}

\makeatother
\usepackage{epstopdf}

\begin{document}

\title{Exploiting Ensemble Learning for Cross-View Isolated Sign Language Recognition
% on Video Swin Transformer
}

\author{Fei Wang}
\email{jiafei127@gmail.com}
\orcid{0009-0004-1142-6434}
\affiliation{%
  \institution{Hefei University of Technology}
  \institution{Institute of Artificial Intelligence, Hefei Comprehensive National Science Center}
  \city{Hefei}
  \country{China}
}

\author{Kun Li}
\email{kunli.hfut@gmail.com}
\orcid{0000-0001-5083-2145}
\affiliation{%
  \institution{CCAI, Zhejiang University}
  \city{Hangzhou}
  \country{China}
}

\author{Yiqi Nie}
\email{nieyiqi@iai.ustc.edu.cn}
\orcid{0009-0001-1739-4515}
\affiliation{%
  \institution{Institute of Artificial Intelligence, Hefei Comprehensive National Science Center}
  \institution{Anhui University}
  \city{Hefei}
  \country{China}
}

\author{Zhangling Duan}
\email{duanzl1024@iai.ustc.edu.cn}
\orcid{0000-0003-3246-8022}
\affiliation{%
  % \institution{Hefei University of Technology}
  \institution{Institute of Artificial Intelligence, Hefei Comprehensive National Science Center}
  \city{Hefei}
  \country{China}
}

\author{Peng Zou}
\email{zonepg666@gmail.com}
\orcid{0009-0000-2452-3023}
\affiliation{%
  \institution{Institute of Artificial Intelligence, Hefei Comprehensive National Science Center}
  \city{Hefei}
  \country{China}
  }

\author{Zhiliang Wu}
\email{wu_zhiliang@zju.edu.cn}
\orcid{0000-0002-6597-8048}
\affiliation{%
  \institution{CCAI, Zhejiang University}
  \city{Hangzhou}
  \country{China}
}

\author{Yuwei Wang}
\email{wyw@ahau.edu.cn}
\orcid{0000-0003-4282-9821}
\affiliation{%
  \institution{Anhui Agricultural University}
  \city{Hefei}
  \country{China}
  }

\author{Yanyan Wei}
\email{weiyy@hfut.edu.cn}
\orcid{0000-0001-8818-6740}
\authornotemark[1]
\affiliation{%
  \institution{Hefei University of Technology}
  \institution{Anhui Province Key Laboratory of Industry Safety and Emergency Technology}
  \city{Hefei}
  \country{China}
  }
\renewcommand{\shortauthors}{Fei Wang et al.}

\begin{abstract}
% In this paper, we present our solution for the Cross-View Isolated Sign Language Recognition Challenge (CV-ISLR) hosted at WWW 2025. 
In this paper, we present our solution to the Cross-View Isolated Sign Language Recognition (CV-ISLR) challenge held at WWW 2025. CV-ISLR addresses a critical issue in traditional Isolated Sign Language Recognition (ISLR), where existing datasets predominantly capture sign language videos from a frontal perspective, while real-world camera angles often vary. To accurately recognize sign language from different viewpoints, models must be capable of understanding gestures from multiple angles, making cross-view recognition challenging. To address this, we explore the advantages of ensemble learning, which enhances model robustness and generalization across diverse views. Our approach, built on a multi-dimensional Video Swin Transformer model, leverages this ensemble strategy to achieve competitive performance. Finally, our solution \textbf{ranked 3rd in both the RGB-based ISLR and RGB-D-based ISLR tracks}, demonstrating the effectiveness in handling the challenges of cross-view recognition. The code is available at: \url{https://github.com/Jiafei127/CV_ISLR_WWW2025}.
\end{abstract}





\maketitle

% https://uq-cvlab.github.io/MM-WLAuslan-Dataset/docs/en/www

\section{Introduction}
%%%%%%%%%%%%%%%%%%%%%%%%%%%%%%%%%%%%%%%%%%%%%%%%%%%%%%%%%%%%%%%%%%%%%%
\begin{figure*}[t]
\centering
\includegraphics[width=1.0\linewidth]{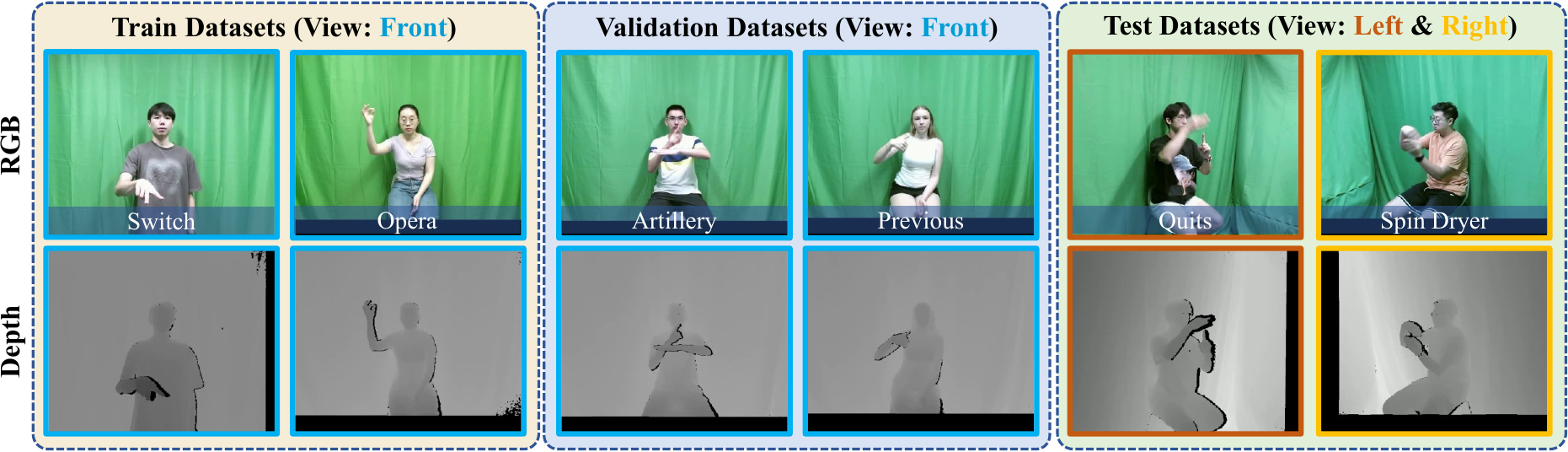}
\caption{The MM-WLAuslan dataset for Cross-View Isolated Sign Language Recognition (CV-ISLR) includes RGB and depth videos in train (front view), validation (left view), and test (left and right views) datasets.}
\label{fig:data}
\end{figure*}
%%%%%%%%%%%%%%%%%%%%%%%%%%%%%%%%%%%%%%%%%%%%%%%%%%%%%%%%%%%%%%%%%%%%%%
\begin{figure}[t]  
\centering
\includegraphics[width=0.9\linewidth]{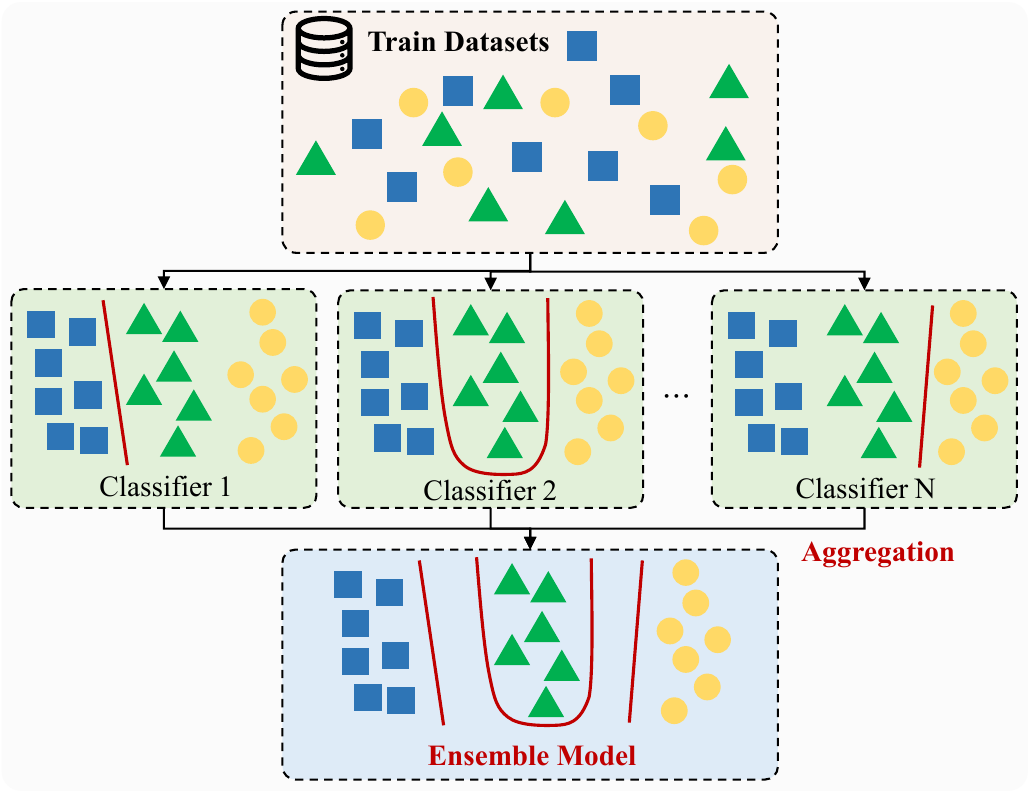}
\caption{Illustration of the ensemble learning process. Multiple classifiers are trained on the same dataset, and their predictions are aggregated to form an ensemble model with improved robustness and accuracy.}
\label{fig:ensemble}
% \vspace{-2.0em}
\end{figure}
%%%%%%%%%%%%%%%%%%%%%%%%%%%%%%%%%%%%%%%%%%%%%%%%%%%%%%%%%%%%%%%%%%%%%%

% Cross-View Isolated Sign Language Recognition (CV-ISLR)~\cite{li2020word,ShenYSDY23} addresses a critical challenge in sign language recognition, where most existing methods assume a fixed frontal view of the signer. However, in real-world scenarios, camera angles often vary significantly, making it difficult for traditional models~\cite{wang2024linguistics,tang2022gloss,tang2021graph,chen2024prototype,liu2024micro,shen2024diverse} to reliably recognize sign language gestures across diverse viewpoints. This challenge is further compounded by the complexity of sign language, which requires precise temporal and spatial understanding of hand and body movements~\cite{shen2024auslan,li2024prototypical,li2024repetitive}, even under occlusions or partial views.
% Therefore, there is a pressing need for robust models that can generalize effectively across multi-view for real-world applicability.
Cross-View Isolated Sign Language Recognition (CV-ISLR)~\cite{li2020word,ShenYSDY23} addresses a critical challenge in sign language recognition, where most existing methods assume a fixed frontal view of the signer. However, in real-world scenarios, camera angles vary significantly, making it difficult for traditional models~\cite{wang2024linguistics,tang2022gloss,tang2021graph,chen2024prototype,liu2024micro,shen2024diverse,li2023joint} to reliably recognize gestures across diverse viewpoints. This challenge is further compounded by the complexity of sign language, requiring precise temporal and spatial understanding of hand and body movements~\cite{shen2024auslan,li2024prototypical,li2024repetitive,zhang2024repetitive}, even under occlusions or partial views. Robust models that generalize effectively across multi-view settings are, therefore, essential for real-world applications.

CV-ISLR involves two main challenges: viewpoint variability and gesture complexity. Viewpoint variability arises from the need to handle diverse camera angles, while gesture complexity stems from the intricate and subtle nature of hand gestures, which can vary significantly even for the same expression~\cite{wei2023mpp,shen2024auslan,sheng2024ai, wei2025leveraging}. To address these challenges, we propose a method that combines the strengths of Ensemble Learning and the advanced Video Swin Transformer (VST)~\cite{liu2021swin,liu2022video,wang2024eulermormer,wang2024frequency}, which has demonstrated strong performance in video understanding tasks. Ensemble Learning~\cite{cao2020ensemble,zhou2022ensemble} is particularly beneficial in CV-ISLR as it aggregates diverse models, improving robustness and generalization across varying viewpoints.

In this work, we explore how ensemble learning can be effectively integrated into the Video Swin Transformer (VST) architecture~\cite{liu2022video} to address the CV-ISLR challenge. Specifically, we conduct extensive experiments using multi-dimensional VST blocks, where we adopt different variants of VST for RGB and Depth inputs—namely, Large, Base, and Small sizes. The idea is to leverage the complementary strengths of these different models at multiple levels of granularity, allowing for better feature extraction and fusion. The final model integrates results from both RGB and Depth streams, producing a comprehensive prediction by aggregating the outputs of both branches through Ensemble Learning.
Overall, we adopt the VST as the baseline, and the main contributions are as follows:
\begin{itemize} 
\item We propose integrating Ensemble Learning into the CV-ISLR framework to improve robustness and generalization, allowing the model to handle viewpoint variability effectively.

\item We investigate the use of multi-dimensional Video Swin Transformer blocks, incorporating different model sizes for both RGB and Depth videos to capture a variety of features at different granularities, enhancing recognition accuracy.

\item We integrate the results from both RGB and Depth streams through Ensemble Learning, ensuring optimal performance by leveraging complementary information from both modalities for more accurate cross-view sign language recognition.
\end{itemize} 

% Our approach significantly enhances performance, as demonstrated by our top-three rankings in both the RGB-based and RGB-D-based ISLR tracks in the CV-ISLR challenge at WWW 2025.

%%%%%%%%%%%%%%%%%%%%%%%%%%%%%%%%%%%%%%%%%%%%%%%%%%%%%%%%%%%%%%%%%%%%%%
\begin{figure*}[t!]
\centering
\includegraphics[width=1.0\linewidth]{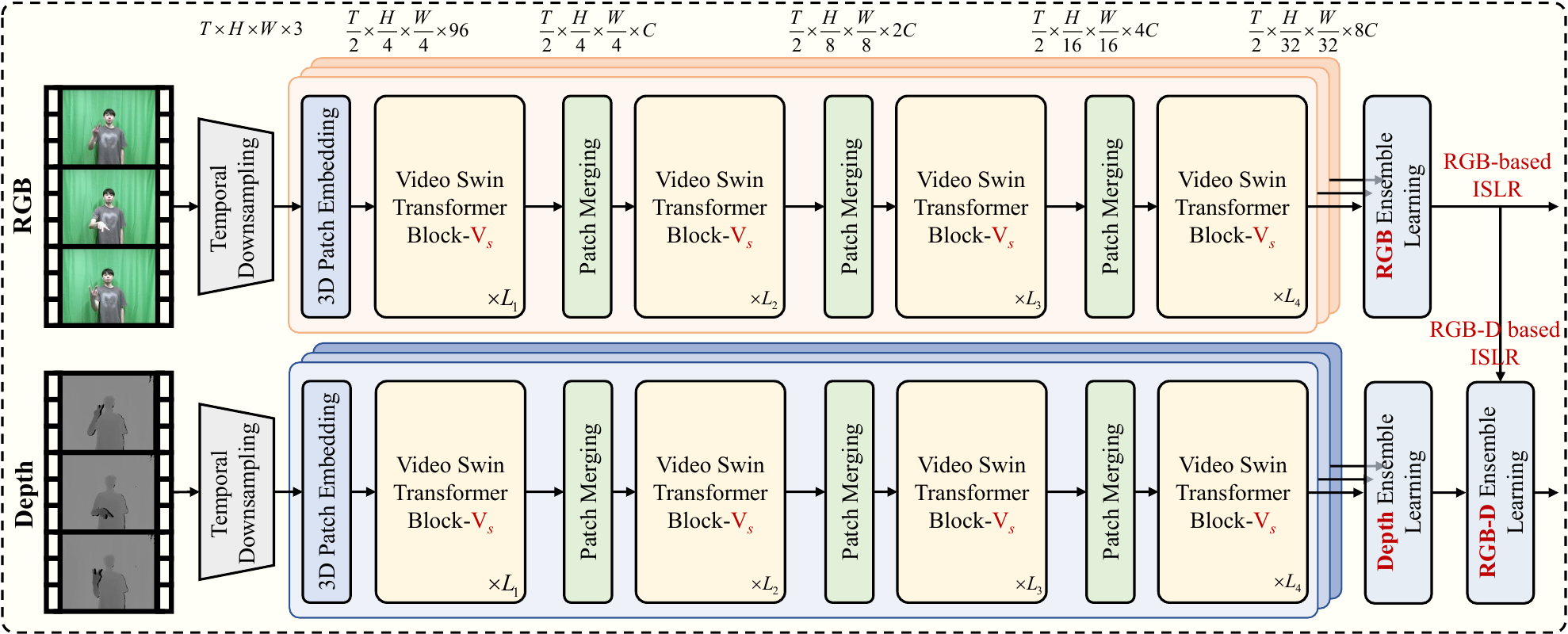}
\caption{Overview of the proposed architecture for CV-ISLR. The architecture processes RGB and depth videos through Video Swin Transformer blocks, with multi-dimensional models for both branches. An ensemble learning approach is applied in two stages: single-modal classification and multi-modal fusion to improve performance across different viewpoints.}
\label{fig:overall}
\end{figure*}

%%%%%%%%%%%%%%%%%%%%%%%%%%%%%%%%%%%%%%%%%%%%%%%%%%%%%%%%%%%%%%%%%%%%%%

\section{Related Work}
\subsection{CV-ISLR Datasets} \label{sec:dataset}
% In Figure~\ref{fig:data}, the MM-WLAuslan dataset~\cite{shen2024mm} is designed to tackle the significant challenges of Cross-View Isolated Sign Language Recognition (CV-ISLR). As the first large-scale, multi-view, multi-modal dataset for Australian Sign Language (Auslan), it contains over 282,000 sign videos, covering 3,215 commonly used Auslan glosses performed by 73 signers.
% Recorded using a multi-camera setup consisting of three Kinect-V2 cameras and one RealSense camera, the dataset captures a diverse range of viewpoints by positioning the cameras hemispherically around the signer, closely simulating the varied angles encountered in real-world scenarios.
% To ensure comprehensive evaluation under practical conditions, the dataset includes four test subsets that introduce challenges such as dynamic environments, background variations, and temporal inconsistencies.
% In this context, our work leverages this dataset to explore new approaches, pushing the boundaries of recognition accuracy across multiple views and modalities.
In Figure~\ref{fig:data}, the MM-WLAuslan dataset~\cite{shen2024mm} tackles the challenges of CV-ISLR as the first large-scale, multi-view, multi-modal dataset for Australian Sign Language (Auslan). It comprises over 282,000 sign videos of 3,215 glosses performed by 73 signers, captured using a multi-camera setup with three Kinect-V2 cameras and one RealSense camera. The hemispherical camera positioning simulates real-world scenarios with diverse viewpoints. To ensure comprehensive evaluation, the dataset includes 4 test subsets addressing dynamic environments, background variations, and temporal inconsistencies. This dataset serves as the foundation for our work, enabling the development of new approaches to improve recognition accuracy across multiple views and modalities.

\subsection{Ensemble Learning} 
Ensemble Learning~\cite{cao2020ensemble,zhou2022ensemble}, which combines multi models to improve performance and robustness, has been widely applied in various recognition tasks.
In Figure~\ref{fig:ensemble}, it effectively reduces bias and variance by aggregating the predictions of diverse models, enhancing accuracy, especially in complex tasks with high variability~\cite{sharma2022trbaggboost}.
Previous works~\cite{cao2020ensemble,yin2024grpose} have leveraged ensemble methods like bagging, boosting, and stacking to combine different neural network architectures and multi-modal inputs, such as RGB, depth, and skeletal data, for more robust models.
Developing an ensemble strategy to integrate models trained in the same view to enhance generalization ability in different views needs to be explored.
% CV-ISLR models benefit from ensemble strategies by integrating models trained on different camera angles, thus enhancing generalization across diverse perspectives.
% Our work extends this idea by integrating Ensemble Learning with Video Swin Transformer models, utilizing multi-dimensional ensembles of varying model sizes to effectively capture features from both RGB and depth inputs, thereby improving performance across cross-view recognition tasks.

\subsection{Video Swin Transformer} 
Swin Transformer~\cite{liu2021swin}, initially developed for image classification, has achieved remarkable success in various vision tasks by capturing local and global dependencies through hierarchical feature extraction. Its extension to video, the Video Swin Transformer~\cite{liu2022video} (VST), effectively models spatiotemporal information and excels in tasks such as action recognition~\cite{li2023data} and video segmentation~\cite{shi2024dust}. For CV-ISLR, VST provides significant advantages by jointly modeling spatial and temporal features. Its hierarchical design captures detailed spatial information and dynamic temporal patterns, while its capability to process multi-modal inputs (RGB and depth) ensures robust feature extraction across diverse views and modalities.

\section{Method}
\subsection{Task Definition} 
We regard the Cross-View Isolated Sign Language Recognition task as a multi-model ensemble prediction problem, with the overall architecture shown in Figure~\ref{fig:overall}. Given input RGB and depth videos, they are downsampled into tensors $\mathbf{Z}_{r} \in \mathbb{R}^{T \times H \times W \times 3}$ and $\mathbf{Z}_{d} \in \mathbb{R}^{T \times H \times W \times 3}$, where $T$, $H$, and $W$ represent the temporal and spatial dimensions. These videos are processed using the Video Swin Transformer (VST) to capture both temporal dependencies and spatial features.
The input is divided into 3D blocks of size $2 \times 4 \times 4 \times 3$, treated as tokens, resulting in 3D token representations of size $\frac{T}{2} \times \frac{H}{4} \times \frac{W}{4}$. Each token is represented by a 96-dimensional feature vector and projected into a higher-dimensional space $C$ through a linear embedding layer, yielding initial feature embeddings $\mathbf{F}_{r} \in \mathbb{R}^{\frac{T}{2} \times \frac{H}{4} \times \frac{W}{4} \times C}$ and $\mathbf{F}_{d} \in \mathbb{R}^{\frac{T}{2} \times \frac{H}{4} \times \frac{W}{4} \times C}$. VST blocks, $\Phi(\cdot)$, further model temporal dynamics and sign language features across four hierarchical stages.
For RGB and depth branches, we implement multi-dimensional VST models (Large, Base, and Small) to capture discriminative features for sign language categories. To optimize performance, we adopt a two-stage ensemble learning: \textbf{(1) Single-Modal Classification Ensemble}, integrating multi-dimensional models within each modality (\eg, RGB video models for RGB-based ISLR); and \textbf{(2) Multi-Modal Fusion Classification Ensemble}, merging RGB and depth branches after their respective integrations to enhance performance in RGB-D-based ISLR tasks.
% The overall prediction is defined as $\mathbf{\Lambda}_{rgb}=\sum_{i=s,b,l}\Phi(\mathbf{Z}_{r};\omega_{rgb})$ for RGB-based ISLR and $\mathbf{\Lambda}_{rgbd}=\sum_{i=s,b,l}\Phi(\mathbf{Z}_{r}\oplus\mathbf{Z}_{d};\omega_{rgbd})$ for RGB-D-based ISLR, where $\mathbf{\Lambda}$ represents the predicted category and $\omega$ denotes the optimized parameters.

\subsection{VST with Multi-Dimensional Models} 
The Video Swin Transformer (VST)~\cite{liu2022video} models both global temporal dependencies and local spatial features through a hierarchical spatiotemporal framework. Its architecture consists of four stages, incorporating 3D-shifted window-based multi-head self-attention mechanisms~\cite{vaswani2017attention,liu2022end} for efficient and localized feature extraction.
Each VST block alternates between regular and shifted 3D window partitioning to capture both intra-window and inter-window dependencies. This is achieved through two key modules: 3D Window-based Multi-Head Self-Attention (3DW-MSA) and 3D Shifted Window-based Multi-Head Self-Attention (3DSW-MSA).
The operations within the $l$-th VST block are expressed as:
\begin{align} 
\hat{z}^{(l)} &= \text{3DW-MSA}(\text{LN}(z^{(l-1)})) + z^{(l-1)}, \\
z^{(l)} &= \text{FFN}(\text{LN}(\hat{z}^{(l)})) + \hat{z}^{(l)}, \\
\hat{z}^{(l+1)} &= \text{3DSW-MSA}(\text{LN}(z^{(l)})) + z^{(l)}, \\
z^{(l+1)} &= \text{FFN}(\text{LN}(\hat{z}^{(l+1)})) + \hat{z}^{(l+1)},
\end{align}
where $\text{LN}$ denotes layer normalization, and $\text{FFN}$ represents a feed-forward network with a two-layer MLP~\cite{tolstikhin2021mlp,li2024patch,zhou2022audio,shen2023adaptive,shen2024spatial} and GELU activation function~\cite{hendrycks2016gaussian,qian2024cluster,zhou2024dense,shen2023mutual}.
The VST’s hierarchical structure progressively downsamples features through patch merging layers, concatenating neighboring tokens, and projecting them into a lower-dimensional space. This enables the model to capture fine-grained spatial details and high-level temporal patterns essential for sign language recognition.
To address the challenges of CV-ISLR, VST implements three model sizes—\textbf{Large}, \textbf{Base}, and \textbf{Small}—for both RGB and depth modalities. Each model is tailored to capture varying levels of discriminative features, providing a comprehensive understanding of sign language gestures across varying perspectives.
% This multidimensional design makes VST highly adaptable to the complex spatiotemporal requirements of CV-ISLR, serving as a robust backbone for both single-modal and multimodal ensemble learning strategies.

\subsection{Ensemble Learning for CV-ISLR}
To leverage the discriminative power of the VST models, we adopt a two-stage ensemble learning strategy tailored for the CV-ISLR.

\subsubsection{Single-Modal Classification Ensemble}
For each modality (RGB or depth), the outputs of the Large, Base, and Small VST models are integrated to produce a robust classification result. \textbf{The RGB branch output is directly used for the RGB-based ISLR task.}
\begin{align} 
\mathbf{\Lambda}_{rgb} &= \sum_{i \in \{s, b, l\}} \Phi(\mathbf{Z}_{r}), \\
\mathbf{\Lambda}_{depth} &= \sum_{i \in \{s, b, l\}} \Phi(\mathbf{Z}_{d}),
\end{align} 
where $\Phi(\cdot)$ denotes the VST model and depth models of size $i$.

\subsubsection{Multi-Modal Fusion Classification Ensemble}
After integrating the RGB and depth models within their respective modalities, \textbf{we perform cross-modal fusion to further enhance recognition performance for the RGB-D-based ISLR task.}
\begin{equation}	  
\begin{aligned} 
\mathbf{\Lambda}_{rgbd} = \sum_{i \in \{s, b, l\}} \Phi(\mathbf{Z}_{r} \oplus \mathbf{Z}_{d}),
\end{aligned}  
\end{equation}
where $\oplus$ denotes the cross-modal integration of the RGB and depth models, and $\omega_{rgbd}^{(i)}$ represents the parameters of the fused RGB-D models.
This strategy effectively combines multidimensional VST models and cross-modal fusion, enabling robust generalization across different viewpoints and modalities.
% Addressing the challenges posed by cross-view and multi-modal data variability significantly improves CV-ISLR's performance.

\section{Experiments}
\subsection{Experiment Setups} 
\noindent\textbf{Datasets.}
Our model is evaluated on the MM-WLAuslan dataset, as mentioned in Section~\ref{sec:dataset}, the first large-scale, multi-view, multi-modal dataset for Australian Sign Language (Auslan).
It comprises over 282,000 sign videos, covering 3,215 commonly used Auslan glosses performed by 73 signers.

\noindent\textbf{Implementation Details.}
During training, $T$, $H$, and $W$ are set to 32, 224, and 224, respectively.
For the VST model, Large, Base and Small sizes of $C$ correspond to 96, 128, and 192, while $\{L_1, L_2, L_3, L_4\}$ = $\{2, 2, 18, 2\}$.
In ensemble learning, the integration ratio of single-modal (RGB or Depth) is $\lambda_l:\lambda_b:\lambda_s$ = $0.4:0.4:0.2$, and the ensemble ratio between multi-modal RGB and depth model is $\lambda_{r}:\lambda_{d}$ = $0.65:0.35$.
Besides, the training process optimizes parameters using cross-entropy loss~\cite{mao2023cross,zhao2024temporal} and the AdamW optimizer~\cite{zhou2024towards}.

\noindent\textbf{Evaluation Metric.} 
We use Top-1 Accuracy (Acc@1)~\cite{zhangtemporal,wang2024low,zhou2022contrastive} as the primary evaluation metric to assess the performance of ISLR models. Top-1 Accuracy measures the proportion of test samples where the model's top prediction matches the ground truth label. 
% This metric provides a clear and straightforward indication of the model's effectiveness in recognizing isolated sign language gestures.

\subsection{Experimental Results} 
\begin{table}[t]
\begin{tabular}{c|c|c}
\toprule
Team & RGB Acc@1 & RGB-D Acc@1 \\ \midrule
VIPL-SLR   &56.87\%& 57.97\%\\
tonicemerald &40.30\%& 33.97\% \\
\rowcolor{gray!20}  % 为这一行添加灰色底纹
\textbf{gkdx2 (Ours)}& 20.29\%& 24.53\% \\ \bottomrule  
\end{tabular}  
\caption{The top-3 result for CV-ISLR on the RGB and RGB-D settings. ``Acc@1'' denotes top-1 accuracy. 
% Data is provided by Codalab Pages~\footnote{The RGB setting page:\url{https://codalab.lisn.upsaclay.fr/competitions/21021#results}}~\footnote{The RGBD setting page: \url{https://codalab.lisn.upsaclay.fr/competitions/21025#results}}
}
\label{tab:top3}
% \vspace{-2.0em}
\end{table}
% \footnotetext[1]{The RGB setting page: \url{https://codalab.lisn.upsaclay.fr/competitions/21021#results}}
% \footnotetext[2]{The RGBD setting page: \url{https://codalab.lisn.upsaclay.fr/competitions/21025#results}}
\begin{table}[t]
\begin{tabular}{l|ccc}
\toprule
Backbone & RGB-based & Depth-based & RGB-D-based    \\ \midrule
VST-Small &   14.84\%  &  14.01\% & - \\
VST-Base &   17.51\%  &  16.46\% & - \\
VST-Large &  17.04\%  &  17.58\%  & -  \\ 
\rowcolor{gray!20}  % 为这一行添加灰色底纹
Ensemble &   \textbf{20.29\%}  & - & \textbf{24.53\%}\\ \bottomrule
\end{tabular}
\caption{Experimental results for RGB and RGB-D tracks on different backbones. 
}
\label{tab:set}
% \vspace{-2.0em}
\end{table}

Table~\ref{tab:top3} summarizes the top-3 results for the CV-ISLR task on RGB and RGB-D tracks. VIPL-SLR achieves the best performance with 56.87\% on RGB and 57.97\% on RGB-D, followed by tonicemerald with 40.30\% and 33.97\%, respectively. Our model (gkdx2) ranks third, achieving 20.29\% on RGB and 24.53\% on RGB-D. These results highlight the robustness of top-performing methods and indicate the potential of our approach, leveraging ensemble learning and Video Swin Transformer, for future improvements in recognition accuracy.
Besides, we report the Top-1 Accuracy (Acc@1) results for different backbones in Table~\ref{tab:set}. It is evident that the ensemble learning strategy effectively captures the strengths of models with different dimensions. In the case of RGB-based ISLR, the ensemble approach achieves a significant performance improvement, with 20.29\% accuracy compared to 17.51\% for the best single model (VST-B).
This performance gain highlights the advantage of leveraging diverse model architectures to enhance robustness and generalization.
% By combining multiple model outputs, ensemble learning mitigates the limitations of individual models and strengthens their complementary capabilities. Furthermore, this strategy proves particularly effective in complex tasks like CV-ISLR, where handling viewpoint variability and multi-modal data is critical. In future work, we aim to further optimize the ensemble design to maximize the synergy between models and achieve even greater recognition accuracy.

% \subsection{Ablation Study} 

\section{Conclusion}
In this work, we address the challenges of CV-ISLR by combining ensemble learning with the VST. By leveraging multi-dimensional VST models across RGB and depth modalities, our ensemble strategy improves robustness and generalization, effectively handling viewpoint variability and gesture complexity. Experiments on the MM-WLAuslan dataset validate the effectiveness of our method, highlighting its potential for advancing robust and generalized solutions for CV-ISLR. In future work, we aim to further optimize the ensemble design to maximize the synergy between models and achieve even greater recognition accuracy.

\begin{acks}
This work was supported by the National Natural Science Foundation of China (Grant No. U24A20331), Anhui Provincial Natural Science Foundation (Grant No. 2408085MF159), the Fundamental Research Funds for the Central Universities (Grant No. PA2024GDSK0096, JZ2023HGQA0472)

\end{acks}
\bibliographystyle{ACM-Reference-Format}
\bibliography{sample-base}
\end{document}